\documentclass[letterpaper, 10 pt, conference]{ieeeconf}  % Comment this line out if you need a4paper

\IEEEoverridecommandlockouts                              % This command is only needed if 
\usepackage{epsfig} % for postscript graphics files
\usepackage{cleveref}
\crefname{figure}{fig.}{fig.}
\crefname{equation}{equ.}{equ.}

\usepackage{url}
\usepackage{multirow}
\usepackage{cite}
\usepackage{threeparttable}
\usepackage{algorithm}
\usepackage{algorithmic}
\usepackage{caption}
\usepackage{subcaption}
\usepackage{siunitx}
\usepackage{color}
\title{\LARGE \bf
Explore-Bench: Data Sets, Metrics and Evaluations for Frontier-based and Deep-reinforcement-learning-based Autonomous Exploration
}

\author{Yuanfan Xu, Jincheng Yu, Jiahao Tang, Jiantao Qiu, Jian Wang,
Yuan Shen, Yu Wang and Huazhong Yang  % <-this % stops a space
\thanks{This work was supported by National Natural Science Foundation of China (No. U19B2019, M-0248), Tsinghua-Meituan Joint Institute for Digital Life, Tsinghua EE Independent Research Project, Beijing National Research Center for Information Science and Technology (BNRist), and Beijing Innovation Center for Future Chips.}
\thanks{All authors are with Department of Electronic Engineering, Tsinghua
University, Beijing, China. \protect\url{xuyf20@mails.tsinghua.edu.cn}, \protect\url{{yu-jc,yu-wang}@tsinghua.edu.cn}}%
}

\begin{document}

\maketitle
\thispagestyle{empty}
\pagestyle{empty}

%%%%%%%%%%%%%%%%%%%%%%%%%%%%%%%%%%%%%%%%%%%%%%%%%%%%%%%%%%%%%%%%%%%%%%%%%%%%%%%%
\begin{abstract}
\label{sec:abstract}
% 机器人自主探索是个重要的任务。为了统一全面地评估探索算法，我们提出了benchmark。该benchmark能够帮助研究者或用户比较和选择探索算法。我们设计了不同的仿真场景（数据集）用来测试探索算法的实用性，提出了评价指标用来评估性能。最后
% 3-level 
Autonomous exploration and mapping of unknown terrains employing single or multiple robots is an essential task in mobile robotics and has therefore been widely investigated. 
Nevertheless, given the lack of unified data sets, metrics, and platforms to evaluate the exploration approaches, we develop an autonomous robot exploration benchmark entitled Explore-Bench. 
The benchmark involves various exploration scenarios and presents two types of quantitative metrics to evaluate exploration efficiency and multi-robot cooperation. 
Explore-Bench is extremely useful as, recently, deep reinforcement learning (DRL) has been widely used for robot exploration tasks and achieved promising results. 
However, training DRL-based approaches requires large data sets, and additionally, current benchmarks rely on realistic simulators with a slow simulation speed, which is not appropriate for training exploration strategies. 
Hence, to support efficient DRL training and comprehensive evaluation, the suggested Explore-Bench designs a 3-level platform with a unified data flow and $12 \times$ speed-up that includes a grid-based simulator for fast evaluation and efficient training, a realistic Gazebo simulator, and a remotely accessible robot testbed for high-accuracy tests in physical environments. 
The practicality of the proposed benchmark is highlighted with the application of one DRL-based and three frontier-based exploration approaches. 
Furthermore, we analyze the performance differences and provide some insights about the selection and design of exploration methods. 
Our benchmark is available at \url{https://github.com/efc-robot/Explore-Bench}.
\end{abstract}

\section{Introduction}
\label{sec:intro}

% 第一段，机器人探索是个重要任务，传统方法多基于frontier，通过贪心或启发式方法挑选frontier
Autonomous exploration of unknown environments aiming to acquire information and to build a map is a fundamental task for mobile robotic systems that has been widely investigated in a single-robot or multi-robot setup. 
Traditional exploration methods are based on frontiers that separate the free from the unknown regions \cite{yamauchi1997frontier, yamauchi1998mobile, burgard2005coordinated, yamauchi1998frontier}. 
The main difference among these strategies considers selecting the frontiers to be visited next, with the majority of these techniques being greedy or heuristic, e.g., cost-based \cite{mei2006energy,osswald2016speeding}, sample-based \cite{umari2017rrt}, and potential field-based exploration \cite{yu2021smmr}.

% 第二段，缺少指标和数据集。
While mobile robot exploration approaches have been widely investigated, there is still a lack of a unified benchmark for performance evaluation. 
The researchers typically challenge their newly proposed exploration strategy against other approaches through custom-designed simulation or experimental scenarios, with the relevant data sets usually being unavailable. 
Moreover, some specific scenarios may be explicitly chosen in favor of their proposed approach. 
Hence, the lack of unified data sets discourages the evaluation from being reproducible and objective. Adding to the unavailable public data sets, there is also a lack of a complete and principled evaluation system appropriate for mobile robot exploration approaches. 
Currently, the performance metric quantifying the exploration results is usually limited to the exploration cost or efficiency, as there is no metric to evaluate the multi-robot cooperation and task allocation, which are essential in multi-robot exploration scenarios.

% 但是，在多机的研究中，多机之间的任务分配和协作也非常重要。我们针对多机的合作提出量化指标。

% Therefore, we design two types of principled evaluation metrics to quantitatively evaluate the exploration efficiency and multi-robot collaboration respectively.

% 除了传统方法，RL逐渐展现出优势。

Deep reinforcement learning (DRL) approaches have been proposed to address the exploration tasks, outperforming traditional approaches in several scenarios \cite{niroui19deep,zhu18deep,Chaplot2020Learning}. 
However, DRL-based methods require iterative interactions with the environment to obtain training data. 
Current robotics benchmarks utilize realistic-but-slow simulation platforms such as Gazebo \cite{koenig2004design}, and Player/Stage \cite{gerkey2003player}, which are incapable of providing sufficient training data for DRL in a reasonable time frame.
% Current robotics benchmarks are realistic but involve processing inefficient simulators such as Gazebo \cite{koenig2004design}, and Player/Stage \cite{gerkey2003player}, which are incapable of providing sufficient training data for DRL in a reasonable time frame. 
Both simulators consume enormous computing resources, as first they render the sensor data and then build the corresponding map. 

Given that DRL algorithms exclude the specific method of location and mapping, we design a grid-based simulator that directly provides the localization and mapping results for DRL-based exploration achieving $12 \times$ speed-up compared to Gazebo. Moreover, DRL-based exploration strategies require diverse training data. Opposing to Gazebo, which provides dozens of scenarios manually, the proposed grid-based simulator can automatically generate thousands of scenarios. 

% Regardless of the proposed fast grid-based simulator, the realistic simulators (such as Gazebo), and the real-world deployment, a comprehensive evaluation of the method tested is mandatory. 
In addition to the fast grid-based simulator, the realistic simulators (such as Gazebo) and the real world deployment are mandatory for comprehensive evaluation.
Therefore, we design a 3-level platform including a fast grid-based simulator, Gazebo, and a real-world testbed. To guarantee the consistency of the evaluation results and reduce the difficulty of deployment on different levels, we employ a unified data flow and interface. 
Overall, we propose an autonomous exploration benchmark called Explore-Bench, providing unified data sets, evaluation metrics and fast-to-deploy platform. We further evaluate the traditional and DRL-based methods on our 3-level benchmark, verify the metrics' effectiveness and practicality, and provide some insights regarding the selection and design of exploration strategies.

The contributions of this paper are summarized as follows:

% a) The grid-based simulator is mainly used to train exploration approaches with reinforcement learning because of its high simulation speed. It can also be used for quick evaluation with potentially slight error.

% b) The realistic simulator Gazebo 

1) \emph{Data Sets}: We design various open-source exploration scenarios to improve the comprehensiveness and objectivity of the evaluation system. Opposing to current benchmarks that involve several fixed simulation scenes, we define the critical elements of common exploration scenarios, i.e., \emph{loop}, \emph{narrow corridor}, \emph{corner}, and \emph{multiple rooms}, and combine them to form rich data sets.

2) \emph{Metrics}: We propose two types of metrics to evaluate the exploration strategies' performance from various aspects.  Specifically, \emph{efficiency metrics}, as a form of quantitative metrics, utilize the total exploration time and the intermediate time when robots cover most unknown regions. \emph{Collaboration metrics} include the standard deviation of the independent exploration areas and the ratio of the overlapping exploration area by multiple robots. 

3) \emph{Platform}: We assist researchers to efficiently develop and evaluate their DRL-based exploration algorithms by designing \emph{a fast grid-based simulator} with $12\times$ speed-up compared to realistic simulators. Combined with \emph{Gazebo} and the \emph{real remotely accessible testbed}, our benchmark forms a 3-level platform.

4) \emph{Evaluations}: We highlight the effectiveness of the proposed benchmark on four representative exploration methods: cost-based frontier exploration \cite{basilico2011exploration}, sample-based RRT-explore \cite{umari2017rrt}, potential field exploration SMMR-explore \cite{yu2021smmr} and DRL-based exploration \cite{Chaplot2020Learning, yu2021surprising}. Each method is implemented in a single-robot setup and a multi-robot one.

\section{Related Work}
\label{sec:relatedwork}

% 写三段 第一段关于机器人探索的基本方法
% 第二段介绍相关的benchmark

Autonomous exploration of unknown environments has been widely investigated in mobile robotics. Traditional methods adopt frontier-based techniques, first proposed by Yamauchi \emph{et al.} \cite{yamauchi1997frontier}. Cost-based methods \cite{mei2006energy, osswald2016speeding, wirth2007exploration} always navigate robots towards the nearest frontier and can be easily extended to a multi-robot setup, where robots share information but select the nearest frontier independently 
\cite{yamauchi1998frontier}. To enable efficient exploration, Umari \emph{et al.} \cite{umari2017rrt} propose a sample-based exploration method that employs multiple Rapidly-exploring Random Trees (RRTs) to detect the frontier. Yu \emph{et al.} \cite{yu2021smmr} design a potential field exploration method termed SMMR-Explore to eliminate the goal’s back-and-forth changes. Deep reinforcement learning (DRL) allows a robot to learn from its own experiences and to generate adaptive goals for unknown environments. Niroui \emph{et al.} \cite{niroui19deep} design an Asynchronous Advantage Actor-critic (A3C) \cite{mnih2016asynchronous} network to output the goal frontier when the map, robot's location, and all possible frontiers are \emph{a priori} known. Zhu \emph{et al.} \cite{zhu18deep} also train an A3C policy, but the output is the next visiting direction. Chaplot \emph{et al.} \cite{Chaplot2020Learning} adopt a global policy and predict a long-term goal for the entire unknown area.

In the context of autonomous robotic exploration, the literature offers several performance evaluations and benchmarking studies. For example, Amigoni \emph{et al.} \cite{amigoni2008experimental} experimentally  assess the strengths and weaknesses of four single-robot exploration strategies. Couceiro \emph{et al.} \cite{couceiro2014benchmark} introduce two performance metrics and conduct several simulation experiments to benchmark five multi-robot exploration and mapping algorithms. Faigl \emph{et al.} \cite{faigl2015benchmarking} develop an evaluation methodology and provide a benchmark to compare frontier-based approaches in a well-defined evaluation environment. Yan \emph{et al.} \cite{yan2015metrics} present a collection of metrics to objectively compare different algorithms that can be applied to collaborative multi-robot exploration. However, none of these works provide public data sets to evaluate exploration approaches nor principled metrics to evaluate each method's exploration efficiency and multi-robot coordination capabilities. Furthermore, the slow simulation speed of the experimental platforms employed does not afford to develop and evaluate DRL-based approaches.

% Furthermore, their experiment platforms are not friendly to the training and evaluation of DRL-based methods.

% In this paper, we present an autonomous exploration benchmark which supports training and evaluation for both traditional and learning-based approaches.
% The benchmark consists of three platforms with different levels of inference speed and realism.
% Public data sets and a principled metric are designed for comprehensive and objective evaluation.
% We also compare the performance of some representative exploration methods on our benchmark and some guidelines for designing algorithms are given.

\section{Data Sets}
\label{sec:datasets}

% 某个探索方法往往在不同的场景下会体现出不同的性能。

% Our data sets contain several scenarios involving both the \emph{basic scenarios} and their combinations.

Our data sets contain the following typical \emph{basic scenarios} and their combinations.
 
% An autonomous exploration approach usually presents different performances when applied to different types of scenarios.
% Since there is a lack of public and comprehensive data sets for training and evaluating mobile robot exploration approaches, we propose and detail our design in this section.
% \textcolor{red}{TO DO: need one sentence to describe how we forms the data sets and we introduce the key elements as follows.}

\begin{figure}[t]
    \centering
    \includegraphics[width=\linewidth]{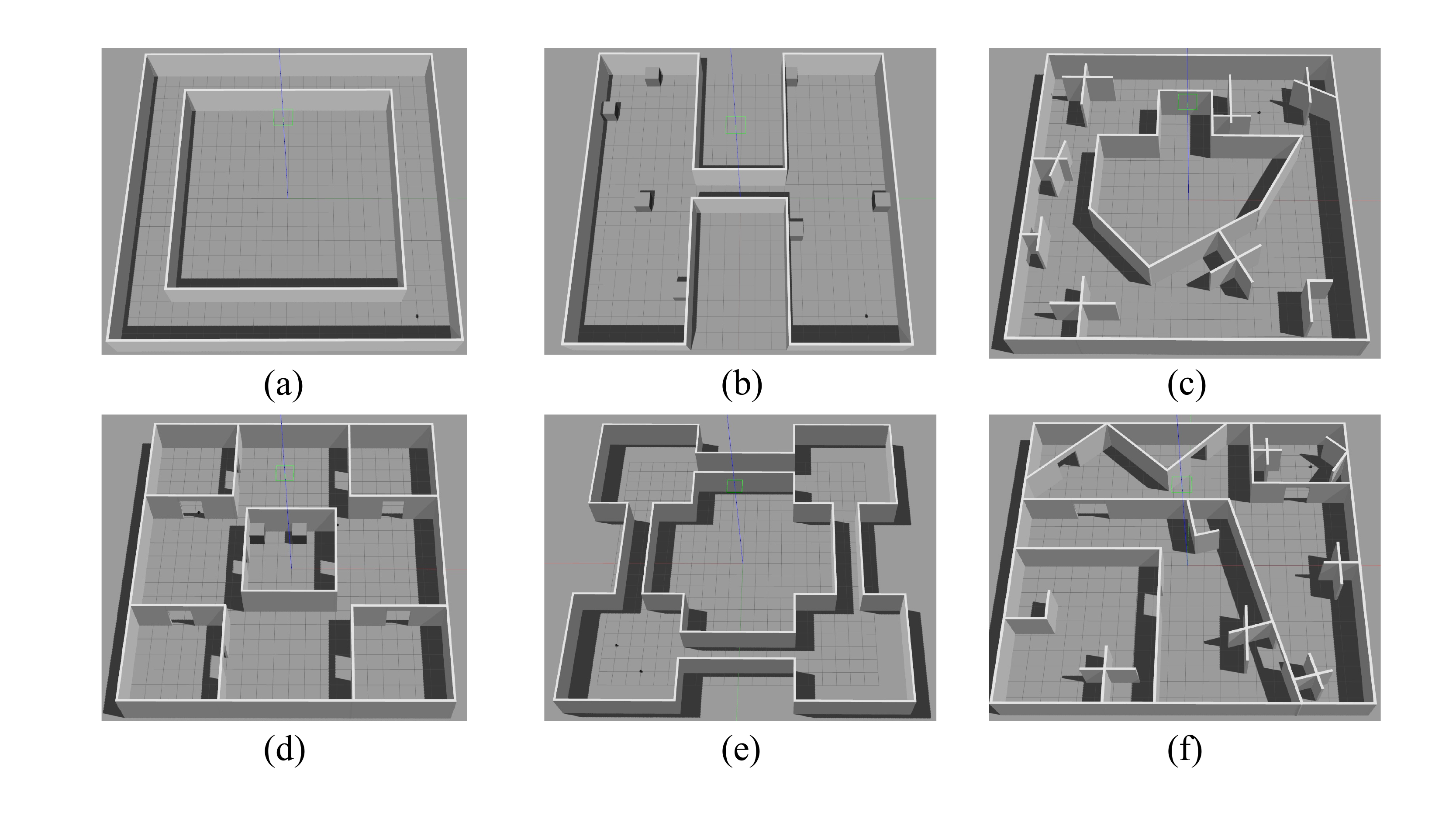}
    \caption{Screenshots of typical simulation scenarios in Gazebo (a) \emph{loop}: square loop, (b) \emph{narrow corridor}, (c) \emph{corner}: clutter environment with many corners, (d) \emph{multiple rooms}, (e) square loop with four narrow corridors, (f) four rooms with many corners.}
    \label{fig:datasets}
    \vspace{-0.6cm}
\end{figure}

\subsection{Loop}

% 回环作为一种环境结构，常出现在各种室内场景中。探索算法在有回环存在的场景中更容易出现backtracking的现象，即revisit map areas that are already explored。所以，它能很好的评估探索算法的稳定和持续性。

A loop often appears in indoor scenarios as an environmental structure, with \Cref{fig:datasets}(a) presenting a simple $20 \times 20m^2$ square loop. 
The shape of a loop can also be circular or other.
Robots are more likely to revisit areas that have already been explored when the task involves a loop closure. This process is known as overlapping and is a significant source for exploration inefficiency. Therefore, the scenarios involving loops are appropriate for evaluating the stability and sustainability of an exploration approach.

\subsection{Narrow Corridor}

% 窄通道常常出现在多个探索区域的连接处，对于机器人而言，选择何时进入窄通道以及进入后能否顺利快速地通过对探索性能有很大的影响。它通常可以用来对探索算法的灵活性和抗干扰能力进行评估。

Narrow corridors often appear as the junction of two regions.
The corridor can be winding or straight, and the width can also be changed.
For example, the scenario illustrated in \Cref{fig:datasets}(b) has a $6m$ long narrow corridor connecting two $7 \times 20m^2$ spaces. When robots come across a narrow corridor, they must decide whether to enter it or to continue exploring the current space. It should be noted that a robot's ability to pass through a corridor smoothly and quickly is essential for autonomous exploration. This scenario is designed to challenge the flexibility and safety of the exploration algorithm under evaluation.

\subsection{Corner}

% 小角落作为一种局部环境特征，代表性上相较于前三者（环境结构）而言略低，但在我们的数据集中将其单独作为一类考虑，是因为它会给绝大多数探索算法造成一定的挑战。小角落是最能体现探索算法在探索完整度和探索效率、短期利益和长期利益之间进行平衡的场景。小角落通常具备的信息量不大，但为了探索完全，机器人往往需要深入或环绕，这时候，选择放弃小角落而去探索其他的大区域能够带来短时间内的最大收益，但是，如果探索的目的是追求完整度，那么机器人在未来势必需要进行折返来覆盖先前未探索的部分，牺牲了长期利益。

% 角落代表了一类信息量不大但为了完整性又不得不进行探索的环境特征
Although corners represent an environmental feature type with limited information gain, they still must be explored for completeness. \Cref{fig:datasets}(c)  presents a typical scenario that is cluttered with small corners. When a robot meets a corner, it can make one of the following decisions. Either pursue the short-term gain, i.e., not exploring the remaining unexplored small corners and prioritize the exploration of large areas, or plan for the long term, i.e., visit every corner in time to avoid backtracking, as these missing unexplored parts might impose a high backtracking cost. This type of scenario can comprehensively reflect the balance between exploration completeness and efficiency and the trade-off between an exploration approach's short and long-term gain.

\subsection{Multiple Rooms}

% 多房间在探索任务中是非常常见的场景，多见于室内环境的探索和建图，主要的区别在于：房间的数量，房间的大小，房间的分布。它可以用来对探索算法的整体性能进行评估。

Multiple rooms are common in indoor exploration tasks, with a typical office-like environment considering five $6 \times 6 m^2$ rooms presented in \Cref{fig:datasets}(d). A specific scenario may vary in the number, size, and distribution of the rooms. In this work, we design scenarios involving multiple rooms to evaluate the overall performance of an exploration method.

\subsection{Combination}

% 单一的环境结构所组成的场景可以对探索算法的某些特性进行评估，这也是大多数现存的探索benchmark所使用的实验环境。但我们认为这还不够，要想全面地评估性能，需要更复杂的场景，在本小节中，我们提出一种设计和构建复杂场景的方法，基于该方法，在Level-1和Level-2中我们精心构建好了一系列数据集，同时因为是开源项目，也欢迎用户根据我们的方法进行扩充，在Level-0中，我们提供一个根据该规则自动生成场景的工具。

% 我们以上所提出的这些场景，虽然结构简单，但能够对探索算法的特定关键属性进行检验，我们称之为“基础场景”。我们将这些基础场景进行组会得到更复杂的场景，从而对探索算法的整体性能进行评估。

The above \emph{basic scenarios} are simple in structure and evaluate specific key properties of the exploration algorithms. Nevertheless, we  combine these and build more complex scenarios to evaluate the overall performance of the exploration approaches. \Cref{fig:datasets}(e)-(f) depict two combination examples: a square loop with four narrow corridors and rooms with many corners. It is worth noting that more combinations are open-source on our website and welcome the contribution from the community to enrich the data sets.

Considering Gazebo and real environment scenarios, these must be created manually. Therefore, we design ten simulation scenarios for Gazebo and five for the real testbed. Our benchmark also aims to provide rich data sets on the grid-based simulator for efficient DRL training, and therefore we provide thousands of scenarios generated automatically on the proposed fast simulator. The generation details of the latter will be introduced in  \Cref{sec:platform}.

\section{Metrics}
\label{sec:metric}

% 结束判定：1. 检测不到新的frontier 2. 探索面积99% 3. 探索时间超出限制
% 评估整体性能：探索时间和覆盖率的曲线面积 多机指标：重复探索率 负载均衡率

\subsection{Efficiency Metrics}

% 在先前的许多探索算法研究或benchmark工作中，
% 机器人探索是一个过程，要想合理并且全面的评价一个探索算法，不仅要关注最终的探索结果，也需要关注探索中的行为。因此，我们选择使用探索率随时间的变化曲线，即Coverage-Time Curve，来评价探索算法。
% 首先，从曲线上，我们可以直接获取到Exploration Time和Ratio这两个对最终结果的评价指标
% 1. 过程和结果同样重要，记录下探索率随时间变化的曲线可以体现出更多元化的信息和特点。 2. 探索算法除了被用于覆盖未知区域，还常被用于目标搜索、搜救、救灾等场景，在这些任务中，人们非常关心机器人对区域的整体情况的快速建模（对受灾地形和大致情况的确认）、早期的探索效率（更快找到目标）等，然而先前常使用的完全覆盖的相关指标无法体现这些能力，因为完全覆盖通常要求机器人在探索过程中注重区域内的所有细节，对此，我们提出在考察完全覆盖的时间外，加入达到90%覆盖率所需时间这一指标。

In current autonomous exploration studies, common performance evaluation metrics are the total time required for exploration and the distance traveled per robot at the end of an exploration run. An exploration run terminates when the target exploration percentage of the entire environment is completed, e.g., 99\%.

Typically, an autonomous exploration task is defined as a process where robots gradually explore unknown regions. Therefore, the robot's behavior during exploration is equally important to the final exploration results, reflecting the algorithm's performance from a new perspective. Hence, two efficiency metrics are proposed as follows:
% We plot the curve of the exploration rate over time and name it as \textbf{Coverage-Time Curve}. 
% The curve serves as an excellent performance metric to demonstrate the entire exploration process.

% From this curve, we can derive some quantitative metrics to facilitate a straightforward comparison. 
\begin{itemize}
    \item The total time $T_{total}$ when the exploration ratio is 99\%.
    \item The time $T_{topo}$ when the exploration ratio is 90\%.
    In most scenarios, a 90\% coverage indicates that the robots have acquired essential \textbf{topological information}, including the overall terrain structure and connectivity, which for exploration tasks such as disaster relief and target search are more important features than map completeness. 
\end{itemize}

Exploiting these two metrics simultaneously reflects the strength and weakness of an exploration algorithm, as low exploration efficiency in the early stage to model all the details produce a small $T_{total}$ but large $T_{topo}$, which does not pose an optimum solution.

\subsection{Collaboration Metrics}

% Coverage-Time Curve provides a complete evaluation of the exploration process.
Our benchmark scheme supports multi-robot systems evaluation, and therefore we design the corresponding robot collaboration metrics to analyze the multi-robot exploration performance quantitatively.  The rationality and fairness of the multi-robot exploration region allocation are evaluated through the load balance metrics. A well-performing multi-robot exploration approach should balance the areas explored by different robots, enhancing the system's efficiency and robustness, and avoid uneven task distribution.

Let  $N$ robots explore and map an unknown environment, with  $S_i, 1 \leq i \leq N$ the area covered by the $i$-th robot when the exploration run ends. In this work, we evaluate the load balance of an exploration method utilizing  the standard deviation $\sigma$ of the independent exploration areas: 
\begin{equation}
    \sigma = \sqrt{\frac{\sum_{i=1}^N (S_i-\overline{S})^2}{N}}
\end{equation}
\begin{equation}
    \overline{S} = \frac{\sum_{i=1}^N S_i}{N}
\end{equation}

Collaborative multi-robot exploration approaches greatly enhance exploration efficiency. Thus we evaluate the collaboration effectiveness by calculating the ratio $r_o$ of the overlapping area $S_o$ explored by multiple robots: 
\begin{equation}
    r_o = \frac{S_o}{S_{total}} = \frac{\sum_{i=1}^N S_i - S_{total}}{S_{total}}
\end{equation}
where $S_{total}$ is the total area of the terrain to be explored.

\section{Platform}
\label{sec:platform}

% In this section, we will describe the 3-level training-evaluation platform in detail: 1) \emph{Level-0}: a grid-based simulator; 2) \emph{Level-1}: a realistic simulator Gazebo; 3) \emph{Level-2}: a real remotely accessible robot testbed.

\begin{figure*}[t]
    \centering
    \includegraphics[width=\linewidth]{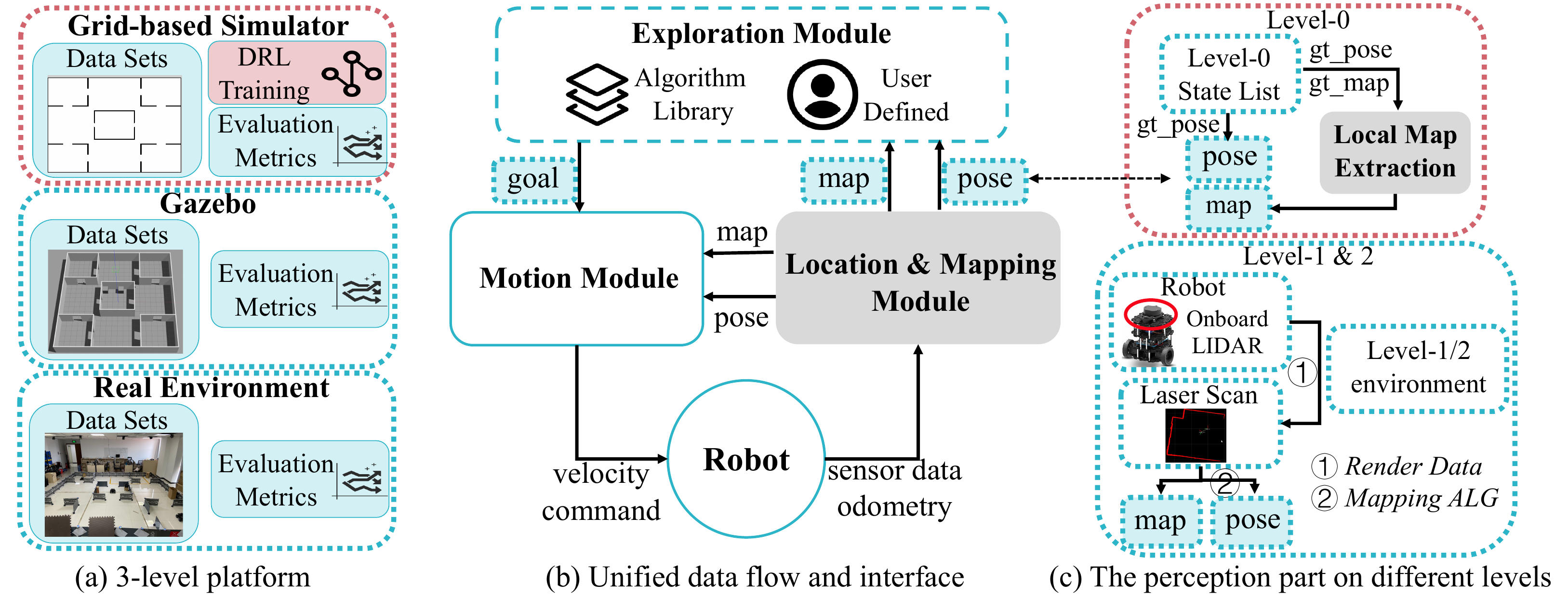}
    \caption{(a) The overview of our proposed 3-level platform, including a grid-based simulator for fast DRL training, Gazebo and real environment for evalution. (b) The unified data flow and interface (adapted and concluded from previous work \cite{umari2017rrt, yan2015metrics}). Different exploration strategies take unified inputs and output the exploration goal. (c) The different perception parts. Level-0 utilizes a new approach to simplify the rendering, location and mapping and accelerates the simulation (\emph{gt} is the abbreviation of ground truth).}
    \label{fig:system}
    \vspace{-0.5cm}
\end{figure*}

To deploy and evaluate various exploration methods on the proposed data sets and metrics, we develop a 3-level platform involving a grid-based simulator (\emph{Level-0}), a realistic simulator Gazebo (\emph{Level-1}), and a real robot testbed (\emph{Level-2}) as shown in \Cref{fig:system}(a). It is worth noting that the \emph{Level-0} simulator is especially designed for fast DRL training.

\subsection{Data flow and interface}

A complete autonomous exploration system follows the unified data flow presented in \Cref{fig:system}(b). The robots initially perceive their surroundings exploiting their onboard sensors, localize themselves and build the corresponding map. In a multi-robot system setup, the individually built maps are shared among all robots and merged into a single unified map. Then the exploration algorithm exploits the merged map and robot pose as input and outputs the goal locations to explore the environment further.

Our benchmark scheme utilizes the Robot Operating System (ROS) \cite{quigley2009ros} and its open-source or modified packages to implement the unified data flow and interface for the \emph{Level-1} and \emph{Level-2} evaluation stages, containing the following modules.

\begin{itemize}
    \item \emph{Location Module}: The ground truth pose originating from the Gazebo parameters is employed for accurate localization. Given that in real environments the ground-truth robot pose is unknown, we utilize the \emph{robot\_pose\_ekf} \cite{robotposeekf} package to estimate its 3D pose.
    \item \emph{Mapping Module}: We adopt \emph{Cartographer} \cite{cartographer} to perform a laser-based SLAM to build a 2-D occupancy grid map. Our platform also supports other laser-based mapping algorithms, such as \emph{gmapping} \cite{gmapping}.
    In multi-robot scenarios, we modify the \emph{multirobot\_map\_merge} \cite{multirobotmapmerge} package to merge the maps created from all robots by exploiting their relative positions. 
    \item \emph{Exploration Module}: This module affords users to design their exploration algorithms and compare in our provided algorithm library. \textbf{Each exploration method utilizes a unified 2D-map and the robots' locations as input and outputs a goal (coordinates of a position) to navigate to. }
    Hence, the proposed 3-level platform provides a unified interface for different exploration methods and different levels. 
    \item \emph{Motion Module}: We adopt the  \emph{move\_base} \cite{movebase} package for the path planning and collision avoidance.
\end{itemize}

\subsection{Fast grid-based simulator (Level-0)}

\emph{Level-1} and \emph{Level-2} are sufficient for evaluating traditional exploration methods. 
However, the expensive data resulting from the slow simulation speed prohibits efficiently training and evaluating deep learning techniques.
% However, the high operating costs of a powerful computer in combination with the slow simulation speed prohibit efficiently training and evaluating deep learning techniques. 
The perception part in the unified data flow consumes huge computing resources and limits the simulation speed, with the most computation-intensive operations being the \emph{Render Data} and \emph{Mapping Algorithm} (\Cref{fig:system}(c)).
The former acquires sensor data from the environment and the latter converts the data into a map and optimizes the robot pose. These two operations impose the highest processing burden during a single processing step.
Therefore, current exploration benchmarks do not consider DRL-based methods. \textbf{To speed up the robots' perception during simulation, we exploit the simplicity and controllability of our grid-based simulator and directly extract the local ground-truth map to quickly provide the inputs for the exploration methods under evaluation.} Concretely, we first intercept the unobstructed perceptible area according to the robot's current location and sensor range and then use Bresenham's line algorithm \cite{bresenham1965algorithm} to simulate the 2D LIDAR scanning and mapping process.
In the multi-robot setup, as a grid map is stored as an array, map merging is achieved by traversing the elements in the array and updating the status.

Considering the remaining modules: 

\emph{Location Module}: The grid-based simulator updates a list with the robots' states and poses.

% \emph{Map Merge}: the grid map is stored as an array in \emph{Level-0}. Hence, map merging is achieved by traversing the elements in the array and updating the status.
\emph{Motion Module}: We provide a global planner utilizing the A* algorithm that aims to find the optimal and collision-free path. Then we calculate the robot's approximate orientation and velocity at each grid cell along the path. Compared with Gazebo and the real robotics testbed, simulating the robot's actual continuous motion can be neglected in the grid-based simulator, i.e., the robot ``teleports" between the grid cells accelerating the simulation. 

% 1) \emph{Map Merge}: we can obtain the relative pose between robots from the grid-based simulator directly as we do in \emph{Location Module}, since the grid-based simulator maintains a list of robots' states, including pose, map and so on. 2) \emph{explore}: this sub-module is the same as that in \emph{Level-1}. 
% 3) \emph{Motion Module}: we provide a global planner with A* algorithm to find the optimal and collision-free path. Then we can  approximately calculate the orientation and velocity of the robot at each grid cell along the path.

% Compared with Gazebo and the real robotics platform, the simulation of the actual continuous movement of the robot can be neglected in the gird-based simulator, that is, the robot can ``teleport" between grid cells, which can also accelerate the simulation. 
% Moreover, for those exploration approaches that output velocity instead of goal, our grid-based simulator also supports calculating the continuous trajectory and converting it to a set of grid cells.

% \begin{figure}[t]
%     \centering
%     \includegraphics[width=\linewidth]{fig/optimized_perception.pdf}
%     \caption{The perception part of different platforms.}
%     \label{fig:perception}
% \end{figure}

\subsection{Diverse simulation scenarios generator}

Diverse training scenarios benefit obtaining a robust and efficient DRL-based exploration policy. Unlike the scenarios in \emph{Level-1/2} that are constructed manually, we present for \emph{Level-0} two methods to create data sets automatically: 

1) Relying on the Gazebo's simulation scenario blueprints. Once designing a scenario in Gazebo is completed, we employ the  \emph{gazebo\_ros\_2Dmap\_plugin} \cite{gazeboros2Dmapplugin} to generate a 2D occupancy map from the simulated world. The grid map from \emph{Level-1} can be directly added to the data sets of \emph{Level-0}.

2) A built-in tool for automatically generating scenarios.
A simulation scenario in the \emph{Level-0} is essentially a ternary matrix (occupied, free and unknown cases). Setting the value of elements in the matrix affords dividing rooms, building narrow corridors and loops, and placing corners. According to our combination of rules presented in \Cref{sec:datasets}, we can generate countless simulation scenarios with little effort.

\section{Experiments And Results}
\label{sec:experiment}

% Level-0和Level-1仿真器运行的设备配置 仿真所用的机器人和传感器 

\subsection{Setup}

As already mentioned, we rely on Gazebo for the \emph{Level-1} simulator and implement the autonomous mapping and exploration utilizing some modified ROS packages. The simulation robot is a Turtlebot3 Burger \cite{turtlebot} with a $360^{\circ}$ laser scanner, whose distance range is $7 m$. The \emph{Level-0} grid-based simulator is implemented in Python and is accelerated through \emph{multiprocessing}. All simulations and evaluations are performed on a Desktop PC with an Intel Core I7-7920 processor and an NVIDIA 1080Ti GPU. Considering the \emph{Level-2} testbed, we employ a real Xtark robot \cite{xtark} equipped with a laser scanner, an IMU sensor, and a wheel odometer.

\subsection{Comparison of inference speed}

% 1m/s 速度 探索10m的直通道。level-2: 10s  level-1: 3.3s level-0: 1.43 s
% 资源使用情况 level-1: 515% CPU level-0: 100% CPU
% 12 倍的 speed up

% 我们通过比较仿真单次探索任务的时间和CPU的占用率来说明Level-0对于快速验证和RL训练的重要性。
% 假设一个机器人需要探索完一条长为10m的直道，速度保持1m/s，在真实的环境中，最理想情况需要10s，Gazebo本身提供了加速仿真的功能，在我们的电脑上，在保证建图和路径规划能正常进行的条件下，最多可加速3倍，即3.3s完成，同时CPU的占用率为515%，但使用我们Level-0仿真器，模拟这样10m长直道的建图和运动只需要1.43s，并且只占用100%的CPU，这意味着在同样的计算资源情况下，我们还可以同时进行5个环境的并行，从而达到相对于真实环境39.6倍的加速。
Considering a single exploration task, we evaluate the simulation time and the CPU usage to illustrate the speed-up of the proposed platform, and demonstrate its appropriateness for fast DRL training and evaluation.

Suppose the scenario where a robot explores a straight corridor of $10 m$ long.
The maximum translational velocity is less than $1m/s$ \cite{turtlebot, xtark}.
Hence, a real robot requires at least $10 s$ to complete this task.
Gazebo can achieve a high simulation speed-up, only accelerating the motion of the robot.
However, \textbf{as processing laser scan and mapping consume lots of computing resources, the largest speed-up factor for Gazebo during exploration is $3 \times$ on our desktop, raising the CPU usage to 515\%} and requiring $3.3 s$ to complete the simulation.
% while Gazebo can accelerate the simulation process. 
% Gazebo在只加速机器人运动的情况下可以达到很高的加速比，但因为处理激光雷达数据和建图耗费了大量计算资源，在我们的电脑上，
% On our desktop, during the exploration and mapping process, the largest speed-up factor for Gazebo is $3 \times$, raising the CPU usage to 515\% and requiring $3.3 s$ to complete the simulation. 
For the same task, our \emph{Level-0} grid-based simulator requires $1.43 s$ imposing a 100\% CPU usage and thus affords parallelizing five more environments under the same computing resources. Hence, the training and evaluation efficiency increases by $12 \times$ compared to Gazebo and $36 \times$ to a real-world scenario as illustrated in \Cref{tab:speed}. 

\begin{table}[t]
\centering
\caption{Speed comparison on different levels}
\label{tab:speed}
\begin{tabular}{|c|c|c|c|}
\hline
        & \multicolumn{1}{l|}{time ($s$)} & \multicolumn{1}{l|}{CPU usage (\%)} & \multicolumn{1}{l|}{speed-up} \\ \hline
Level-0 & 1.43                          & 100                                 & 36x                         \\ \hline
Level-1 & 3.3                           & 515                                 & 3x                          \\ \hline
Level-2 & 10                            & -                                   & 1x                            \\ \hline
\end{tabular}
\vspace{-0.5cm}
\end{table}

\subsection{Consistency of evaluation results}

% detail实验细节：我们采用的是immediate replanning的方式，最符合实际探索所使用的方法，同时也cover本身决策算法所消耗的时间，因为我们的计算资源是同样的，对比是公平的。讲move_base参数的一致性，最大线速度和角速度。指标是3次平均取均值。
We also evaluate the consistency of the experimental results obtained from our benchmark by deploying two exploration approaches in the five rooms scenario illustrated in \Cref{fig:datasets}(d). 
% Since the robot motion and perception accuracy on different levels are different, we focus on the order of performance.
\Cref{tab:consistency} indicates that the order (\emph{field} performs better than \emph{cost}) is preserved on different levels and the robot travels along a similar trajectory.
Hence, the evaluation results remain consistent, highlighting that simulation acceleration through our grid-based simulator is effective.
% \Cref{tab:consistency} indicates a slight error in the quantitative statistics and that the order (\emph{field} performs better than \emph{cost}) is preserved on different levels. Hence, the evaluation results remain consistent, highlighting that simulation acceleration through our grid-based simulator is effective.

\begin{table}[t]
\centering
\caption{Consistency evaluation for different levels}
\label{tab:consistency}
\begin{tabular}{|p{8.7mm}<{\centering}|p{4mm}<{\centering}|c|p{4mm}<{\centering}|c|p{4mm}<{\centering}|c|p{4mm}<{\centering}|c|}
\hline
\multirow{2}{*}{} & \multicolumn{2}{c|}{$T_{topo}$ ($s$)} & \multicolumn{2}{c|}{$T_{total}$ ($s$)} & \multicolumn{2}{c|}{$\sigma$ ($m^2$)} & \multicolumn{2}{c|}{$r_o$ (\%)} \\ \cline{2-9} 
                  & cost        & field       & cost         & field         & cost       & field       & cost         & field         \\ \hline
Level-0           &  134           &      \textbf{100}       &  172            &     \textbf{125}    &     0.09       & \textbf{0.01}   &      0.10        &  \textbf{0.04}         \\ \hline
Level-1           &       145      &     \textbf{101}        &        201      &      \textbf{131}         &     0.08       &        \textbf{0.04}     &      0.32        &      \textbf{0.16}         \\ \hline
Level-2           &      119       &      \textbf{84}       &        130      &       \textbf{93}        &     0.08       &     \textbf{0.01}        &       0.21       &      \textbf{0.16}         \\ \hline
\end{tabular}
\vspace{-0.5cm}
\end{table}

\subsection{Evaluation}

% 1) cost-based  2) utility-based  3) RL-based  4) Sample-based  5) Field-based 

\begin{table*}[]
\centering
\caption{Performance evaluation of exploration approaches in single-robot scenarios.}
\label{tab:single_robot}
\begin{tabular}{|c|c|c|c|c|c|c|c|c|}
\hline
\multirow{2}{*}{} & \multicolumn{2}{c|}{cost}   & \multicolumn{2}{c|}{sample} & \multicolumn{2}{c|}{field}  & \multicolumn{2}{c|}{DRL} \\ \cline{2-9} 
                  & $T_{topo} $ ($s$)         & $T_{total}$ ($s$)      & $T_{topo}$ ($s$)         & $T_{total}$ ($s$)   & $T_{topo}$ ($s$)         & $T_{total}$ ($s$)      & $T_{topo}$ ($s$)         & $T_{total}$ ($s$)      \\ \hline
loop              & \textbf{124} & \textbf{145} & 145             & 180       & 131          & 152          & 190       & 214          \\ \hline
corridor          & 162          & 169          & 166             & 170       & \textbf{158} & \textbf{162} & 160       & 266          \\ \hline
corner            & 210          & 426          & 171             & 331       & \textbf{133} & \textbf{324} & 324       & 381          \\ \hline
room              & 159          & 210          & 176             & 211       & \textbf{156} & \textbf{191} & 270       & 315          \\ \hline
comb1             & 169          & \textbf{175} & \textbf{141}    & 192       & 165          & 183          & 249       & 297          \\ \hline
comb2             & 230          & \textbf{537} & 249             & 439       & \textbf{224} & 547          & 588       & 755          \\ \hline
\end{tabular}
\vspace{-0.2cm}
\end{table*}

\begin{table*}[]
\centering
\caption{Performance evaluation of exploration approaches in multi-robot scenarios.}
\label{tab:multi_robot}
\begin{tabular}{|p{8mm}<{\centering}|c|p{6mm}<{\centering}|p{6mm}<{\centering}|p{5mm}<{\centering}|p{4mm}<{\centering}|p{6mm}<{\centering}|p{6mm}<{\centering}|p{5mm}<{\centering}|p{4mm}<{\centering}|p{6mm}<{\centering}|p{6mm}<{\centering}|p{5mm}<{\centering}|p{4mm}<{\centering}|p{6mm}<{\centering}|p{6mm}<{\centering}|p{5mm}<{\centering}|p{4mm}<{\centering}|}
\hline
\multicolumn{2}{|c|}{\multirow{2}{*}{}} & \multicolumn{4}{c|}{cost}                                                                                                                                                                                              & \multicolumn{4}{c|}{sample}                                                                                                                                                                                            & \multicolumn{4}{c|}{field}                                                                                                                                                                                             & \multicolumn{4}{c|}{DRL}                                                                                                                                                                                                \\ \cline{3-18} 
\multicolumn{2}{|c|}{}                  & \begin{tabular}[c]{@{}c@{}}$T_{topo}$\\ $(s)$\end{tabular} & \begin{tabular}[c]{@{}c@{}}$T_{total}$\\ $(s)$\end{tabular} & \begin{tabular}[c]{@{}c@{}}$\sigma$\\ $(m^2)$\end{tabular} & \begin{tabular}[c]{@{}c@{}}$r_o$\\ $(\%)$\end{tabular} & \begin{tabular}[c]{@{}c@{}}$T_{topo}$\\ $(s)$\end{tabular} & \begin{tabular}[c]{@{}c@{}}$T_{total}$\\ $(s)$\end{tabular} & \begin{tabular}[c]{@{}c@{}}$\sigma$\\ $(m^2)$\end{tabular} & \begin{tabular}[c]{@{}c@{}}$r_o$\\ $(\%)$\end{tabular} & \begin{tabular}[c]{@{}c@{}}$T_{topo}$\\ $(s)$\end{tabular} & \begin{tabular}[c]{@{}c@{}}$T_{total}$\\ $(s)$\end{tabular} & \begin{tabular}[c]{@{}c@{}}$\sigma$\\ $(m^2)$\end{tabular} & \begin{tabular}[c]{@{}c@{}}$r_o$\\ $(\%)$\end{tabular} & \begin{tabular}[c]{@{}c@{}}$T_{topo}$\\ $(s)$\end{tabular} & \begin{tabular}[c]{@{}c@{}}$T_{total}$\\ $(s)$\end{tabular} & \begin{tabular}[c]{@{}c@{}}$\sigma$\\ $(m^2)$\end{tabular} & \begin{tabular}[c]{@{}c@{}}$r_o$\\ $(\%)$\end{tabular} \\ \hline
\multirow{2}{*}{loop}          & f$^*$      & 84                                                 & 91                                                    & 0.04                                              & 0.21                                                  & 69                                                 & 85                                                    & 0.12                                              & 0.18                                                  & \textbf{60}                                        & \textbf{74}                                           & \textbf{0.03}                                     & \textbf{0.13}                                         & 130                                                & 190                                                   & 0.11                                              & 0.15                                                  \\ \cline{2-18} 
                               & c$^{\S}$      & 111                                                 & 131                                                    & \textbf{0.03}                                              & 0.91                                                  & 80                                        & 84                                           & 0.06                                              & \textbf{0.13}                                        & \textbf{68}                                                 & \textbf{73}                                                   & 0.05                                    & 0.15                                                 & 94                                                 & 105                                                   & 0.08                                              & 0.15                                                 \\ \hline
\multirow{2}{*}{corridor}      & f      & 51                                                 & 74                                                    & 0.02                                              & 0.06                                                  & 66                                                 & 88                                                    & 0.01                                              & 0.01                                        & \textbf{50}                                        & \textbf{54}                                           & \textbf{0.01}                                    & \textbf{0.01}                                                  & 152                                                & 167                                                   & 0.03                                             & 0.31                                                  \\ \cline{2-18} 
                               & c      & 124                                                & 130                                                   & 0.08                                             & 0.46                                                 & 121                                                & 126                                                   & \textbf{0.01}                                    & 0.70                                                 & \textbf{102}                                       & \textbf{106}                                          & 0.06                                              & \textbf{0.43}                                         & 166                                                & 230                                                   & 0.27                                              & 0.46                                                  \\ \hline
\multirow{2}{*}{corner}        & f      & 168                                                & 185                                                   & 0.07                                             & 0.72                                                 & 173                                                & 233                                                   & 0.36                                              & 0.23                                                  & \textbf{157}                                       & \textbf{172}                                          & \textbf{0.01}                                   & \textbf{0.13}                                        & 406                                                & 496                                                   & 0.27                                              & 0.39                                                 \\ \cline{2-18} 
                               & c      & 174                                                & 185                                                   & 0.13                                             & 0.54                                                  & \textbf{101}                                       & \textbf{132}                                          & \textbf{0.03}                                    & \textbf{0.21}                                        & 129                                                & 152                                                   & 0.11                                              & 0.67                                                & 188                                                & 266                                                   & 0.09                                             & 0.26                                                 \\ \hline
\multirow{2}{*}{room}          & f      & 145                                       & 201                                                   & 0.08                                               & 0.32                                                  & 113                                                & \textbf{138}                                          & 0.13                                             & 0.32                                                 & \textbf{101}                                                & \textbf{131}                                                   & \textbf{0.04}                                    & 0.16                                                 & 249                                                & 407                                                   & 0.09                                             & \textbf{0.16}                                        \\ \cline{2-18} 
                               & c      & 200                                                & 240                                                   & 0.10                                             & \textbf{0.20}                                        & 125                                                & \textbf{135}                                          & 0.18                                              & 0.22                                                  & \textbf{108}                                       & 194                                                   & 0.10                                             & 0.33                                                  & 252                                                & 428                                                   & \textbf{0.08}                                    & 0.34                                                 \\ \hline
\multirow{2}{*}{comb1}         & f      & 156                                                & 167                                                   & 0.03                                             & 0.58                                                 & 202                                                & 213                                                   & 0.02                                             & 0.03                                                & \textbf{94}                                        & \textbf{116}                                          & \textbf{0.02}                                   & \textbf{0.01}                                       & 277                                                & 281                                                   & 0.04                                             & 0.47                                                  \\ \cline{2-18} 
                               & c      & 246                                                & 260                                                   & 0.10                                               & 0.64                                                  & 118                                                & 122                                                   & 0.10                                               & \textbf{0.23}                                         & \textbf{110}                                       & \textbf{116}                                          & \textbf{0.02}                                     & 0.27                                                 & 245                                                & 282                                                   & 0.09                                             & 0.26                                                  \\ \hline
\multirow{2}{*}{comb2}         & f      & \textbf{120}                                       & 320                                                   & \textbf{0.07}                                    & 0.54                                                 & 189                                                & 427                                                   & 0.24                                              & 0.36                                                  & 151                                                & \textbf{248}                                          & 0.09                                             & 0.49                                                 & 291                                                & 617                                                   & 0.17                                             & \textbf{0.22}                                        \\ \cline{2-18} 
                               & c      & 146                                                & \textbf{168}                                          & 0.05                                             & \textbf{0.19}                                        & \textbf{144}                                       & 260                                                   & \textbf{0.03}                                    & 0.34                                                 & 154                                                & 189                                                   & 0.08                                             & 0.54                                                 & 270                                                & 346                                                   & 0.18                                             & 0.29                                                 \\ \hline
\end{tabular}
\begin{tablenotes}
\centering
\item[1] $^{*}$Robots are far away from each other. $^{\S}$Robots are close to each other.
\end{tablenotes}
\vspace{-0.5cm}
\end{table*}

% 介绍方法：
We implement four typical autonomous exploration methods on our benchmark, with each technique implemented in a single and a multi-robot (two robots) version. The challenged methods are:

1) \emph{Cost}: This approach always selects the nearest frontier 
during exploration \cite{yamauchi1997frontier}. For the multi-robot version, robots share perceptual information but select the nearest frontier independently \cite{yamauchi1998frontier}.

2) \emph{Sample}: This method detects frontiers utilizing the rapidly-exploring randomized trees (RRT) \cite{umari2017rrt}. In this work, we extend RRT to facilitate multi-robot scenarios by growing multiple trees independently. 

3) \emph{Potential Field}: 
This approach builds the potential field from the surroundings and selects the frontier with the largest
potential gradient \cite{yu2021smmr}.
% The potential field is built from the surrounding ones by selecting the frontier with the most significant potential gradient \cite{yu2021smmr}. 
Multi-robot cooperation is enabled by introducing a new repulsive potential between the robots.

4) \emph{Deep Reinforcement Learning}: This technique deploys a DRL-based policy to determine the global goal \cite{Chaplot2020Learning} and then the nearest frontier to the global goal is selected. Training the policy for multiple robots requires exploiting the shared information through the entire robotic network, including the merged map and the robot's relative pose. In this work, we only train the policy on the \emph{Level-0} simulator and evaluate the performance on all levels.

\Cref{tab:single_robot} and \Cref{tab:multi_robot} enumerate the quantitative statistics of the exploration approaches in \emph{Level-1} Gazebo for a single-robot and a multi-robot system, respectively. 
The data sets for evaluation are the six exploration scenarios depicted in \Cref{fig:datasets}. 
For simplification, we use \emph{loop}, \emph{corridor}, \emph{corner}, \emph{room}, \emph{comb1 \& 2} to represent them.
The evaluation results in \emph{Level-0} and \emph{Level-2} are not shown in our paper due to the consistency of our platform. 
Given that the robots' start pose affects the exploration performance, especially in the multi-robot system, we study two extreme cases: robots being far away from each other and being close to each other.
The parameters of all modules except exploration module keep the same and researchers can refer to our website for details.
% During the experiments, we observe that the \emph{move\_base}  sometimes freezes, causing inaccurate time measurements. Thus, for a fair comparison, we collect the traveled distance and the velocity and then calculate $T_{90}$ and $T_{total}$.

\subsubsection{Comparison on Efficiency}

% Efficiency上来看，我们认为势场法>RRT>cost>RL，原因是势场法首先考虑了全局信息，势场本身融合了距离和可获得的信息量，RRT法并没有考虑全局信息，frontier的检测只取决于随机树的生成，虽然在选取frontier的时候考虑了距离和信息量，但理论上最优的frontier很可能没有被检测到，因此性能稍逊于势场。cost的方法考虑了全局信息，但是对于frontier的选取只凭借距离，因此性能更差。对于RL，其一般的表现是可以理解的，因为我们所训练的只是一个高层的规划器，没有训练与之配套的frontier检测器和局部规划器，所以会出现高层决策正确，但是机器人具体的运动规划出现偏差，导致反复的情况出现。

The  $T_{topo}$ and $T_{total}$ metrics of \Cref{tab:single_robot} and \Cref{tab:multi_robot} reveal an exploration efficiency ranking from high to low as follows: Potential field, sample, cost and DRL. The potential field method exploits the global information and integrates the distance to the frontier and the potential information gain. 
The sample-based method does not utilize global information and the frontier detection only relies on random tree generation. 
Although the frontier selection depends on distance and information gain, the optimal frontier may not be detected. 
Therefore, the performance of the sample-based approach is inferior to the potential field approach. 
The cost-based method detects all possible frontiers but the selection solely relies on distance. 
Finally, DRL performs worst, as we only train a high-level planner. Without training the corresponding frontier detector and local planner,  robots often misinterpret the high-level decisions leading to a repeated back-and-forth exploration motion.

% 同时，我们能够发现，T_90作为T_total的一个辅助指标，也能揭示很多问题。我们发现，在corner或者room这些具有很多独立小区域的场景中，时常会出现T_90小但T_total大的情况，原因就是不同算法对于优先探索区域的决策不同。如果用户希望尽快掌握整体区域结构而不在乎细节，那应该选择T_90更小的算法。

Our results reveal that both $T_{topo}$ and $T_{total}$ can give significant insights. 
The case where $T_{topo}$ is much smaller than $T_{total}$ occurs in scenarios involving many independent small regions, i.e., multiple rooms and corners, because robots miss some small areas inevitably during exploration and backtracking consumes much time. 
% each exploration approach has different priorities in exploring the unknown area along the original trajectory. 
If users wish to obtain the overall terrain structure as soon as possible without mapping the details, an exploration algorithm with a smaller $T_{topo}$ should be chosen.

% 总之，对于设计一个高效的探索算法，需要具备以下要素：任务的合理分层和分工，即完整地考虑全局和局部的规划器，以及机器人的运动模型；尽量使用全局信息；对于目标点的选择需要同时考虑距离和信息量的约束。

In summary, the following guidelines are helpful when designing an efficient exploration algorithm.
First, utilize global information as much as possible. Then design a global planner for frontier selection and an appropriate local planner to respond to the high-level decision. Finally, define a good trade-off between travel cost and potential information gain.

\subsubsection{Comparison on Cooperation}

% 对于多机探索算法合作性能的评估，我们重点关注sigma和So。我们发现，势场法和RL方法能取得比较好的机器人协作，即负载较为均衡且重复探索的区域面积小，这是因为这两种方法有着针对协作的特别设计，势场法通过引入斥力让机器人尽可能的分开探索，而MARL方法通过对共享信息的特征提取和分析，给出了让机器人分开探索的高层决策，虽然整体的efficiency仍然不高，但合作有效。相比较而言，cost和rrt的方法只能在特定的一些场景下取得好的合作，这是因为在算法设计上机器人就是自私的，一旦没有一个中心节点为它们分配目标，它们就只按自己的节奏进行探索。

To evaluate the cooperative performance of multi-robot exploration approaches, we focus on $\sigma$ and $r_o$ presented in \Cref{tab:multi_robot}. We observe that the potential field and DRL-based methods achieve better collaboration results, affording a balanced travel load and a smaller overlapping area. This is due to the special design of these two methods enhancing multi-robot cooperation, i.e., when two robots move close to each other, the repulsive force drives them to separate. Since we train the policy for multi-robot exploration by employing the Multi-Agent Reinforcement Learning (MARL) scheme, the neural network can extract useful features from the shared information and make high-level decisions to assign robots to different regions. Although the overall efficiency is still not high for the DRL-based approach, the cooperation is effective. In comparison, the cost and sample-based methods achieve good cooperation only in specific scenarios, as these algorithms do not explicitly involve a coordination policy.

Therefore, the core guideline for designing a cooperative multi-robot exploration approach is to effectively utilize the shared information between robots, such as the merged map and relative pose, through heuristic rules or properly trained neural networks.

\section{Conclusion \& Future work}
\label{sec:conclusion}
% Current deep-reinforcement-learning-based exploration approaches challenge the simulation speed of existing robotics simulators and the diversity of data sets. Spurred by the low simulation speed-up and the inefficient evaluation data sets, this work proposes an autonomous exploration benchmark called Explore-Bench. The proposed benchmark involves a 3-level training-evaluation platform through a unified data flow and interface, including the fast grid-based simulator, the more realistic  Gazebo simulator, and the real robotics testbed. We innovatively design the grid-based simulator to quickly train DRL-based exploration strategies achieving $12\times$ speed-up than Gazebo and provide more diverse simulation scenarios. Additionally, we also propose a set of collaboration metrics to evaluate the load balance and cooperation effectiveness. To demonstrate the practicality of the suggested benchmark, we evaluate three traditional frontier-based and one DRL-based exploration method on the Explore-Bench and present some guidelines about the selection and design of exploration strategies. 

In this paper, we propose an autonomous exploration benchmark called Explore-Bench.
To comprehensively evaluate frontier-based and deep-reinforcement-learning-based autonomous exploration approaches, Explore-Bench provides 1) diverse data sets, 2) efficiency and collaboration metrics, and 3) a $12\times$ speed-up grid-based simulator to quickly train and evaluate exploration strategies.
With the application of one DRL-based and three frontier-based exploration approaches on the benchmark, we show that the potential field-based method achieves the best all-around performance and the DRL-based method needs a more elaborate design to outperform traditional methods. 
We also find that the trade-off between travel cost and potential information gain affects exploration efficiency significantly, and the proper use of shared information between robots results in good cooperation during exploration.

As future work, we would like to design additional varied exploration scenarios, including field and subterranean environments, to enrich our data sets and evaluate more DRL-based exploration approaches on our benchmark.

% Various simulation scenarios are elaborately designed
% and principled efficiency and collaboration metrics are proposed.
% We present the application of the proposed benchmark in four typical exploration approaches to demonstrate the practicality of the benchmark.

% In the future, we want to rewrite the \emph{Level-0} simulator in C/C++ to achieve a higher speedup and enhance the data sets and algorithm supports through the open-source community.

\end{document}